\documentclass[runningheads]{llncs}
\usepackage{amsmath,amssymb}
\usepackage[T1]{fontenc}
\usepackage{graphicx}
\usepackage{multirow}
\usepackage{enumerate}

\begin{document}
\title{Learning Large Margin Sparse Embeddings for Open Set Medical Diagnosis} 
%
\author{Mingyuan Liu\inst{1} \and Lu Xu\inst{1} \and Jicong Zhang\inst{1,2,*}}
\authorrunning{Mingyuan Liu, Lu Xu, Jicong Zhang.}

\institute{$^{1}$School of Biological Science and Medical Engineering, \\ Beihang University, Beijing, China\\$^{2}$Hefei Innovation Research Institute, Beihang University, Hefei, Anhui, China
\email{$\{$liumingyuan95, xulu181221, jicongzhang$\}$@buaa.edu.cn}}

%
%
%
\maketitle              
\begin{abstract}

Fueled by deep learning, computer-aided diagnosis achieves huge advances. 
However, out of controlled lab environments, algorithms could face multiple challenges. 
Open set recognition (OSR), as an important one, states that categories unseen in training could appear in testing. 
In medical fields, it could derive from incompletely collected training datasets and the constantly emerging new or rare diseases. 
OSR requires an algorithm to not only correctly classify known classes, but also recognize unknown classes and forward them to experts for further diagnosis. 
To tackle OSR, we assume that known classes could densely occupy small parts of the embedding space and the remaining sparse regions could be recognized as unknowns. 
Following it, we propose Open Margin Cosine Loss (OMCL) unifying two mechanisms. 
The former, called Margin Loss with Adaptive Scale (MLAS), introduces angular margin for reinforcing intra-class compactness and inter-class separability, together with an adaptive scaling factor to strengthen the generalization capacity. 
The latter, called Open-Space Suppression (OSS), opens the classifier by recognizing sparse embedding space as unknowns using proposed feature space descriptors.
Besides, since medical OSR is still a nascent field, two publicly available benchmark datasets are proposed for comparison. 
Extensive ablation studies and feature visualization demonstrate the effectiveness of each design. 
Compared with state-of-the-art methods, MLAS achieves superior performances, measured by ACC, AUROC, and OSCR.

\keywords{Open set recognition \and Computer aided diagnosis \and Image classification.} 
\end{abstract}
\section{Introduction and Related Work}
Deep learning achieves great success in image-based disease classification. 
However, the computer-aided diagnosis is far from being solved when considering various requirements in real-world applications. As an important one, open set recognition (OSR) specifies that diseases unseen in training could appear in testing \cite{osr13}. It is practical in the medical field, caused by the difficulties of collecting a training dataset exhausting all diseases, and by the unpredictably appearing new or rare diseases. As a result, an OSR-informed model should not only accurately recognize known diseases but also detect unknowns and report them. Clinically, these models help construct trustworthy computer-aided systems. By forwarding unseen diseases to experts, not only the misdiagnosis of rare diseases could be avoided, but an early warning of a new disease outbreak could be raised.

\begin{figure}[t]
\includegraphics[width=\textwidth]{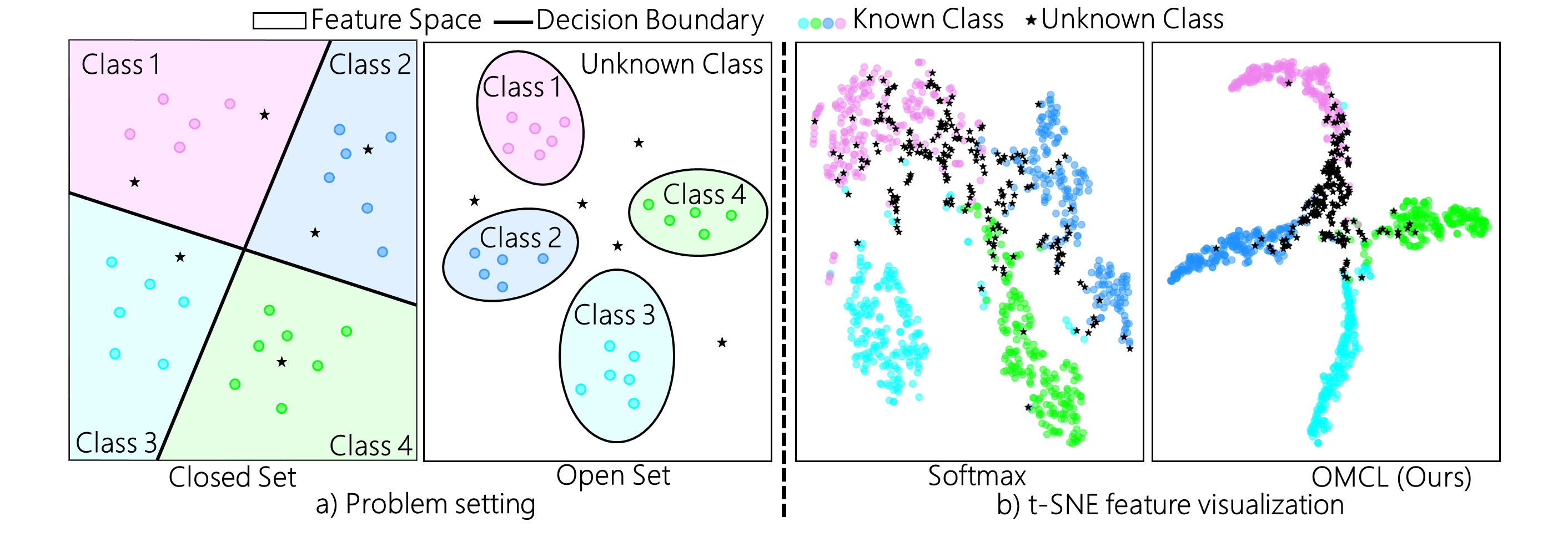}
\caption{a) Diagrams of the open set recognition problem, and b) the feature visualization of original closed set classifier and our proposed OMCL.} 
\label{fig:intro}
\end{figure}

There are many fields related to OSR but are essentially different. 
In classification with reject options \cite{rej1,rej2},  samples with low confidence are rejected to avoid misclassification. However, since its closed set nature, unknown classes could still be misclassified confidently \cite{osr13,review1}. 
Anomaly detection, novelty detection, and one-class classification\cite{ad} aim at recognizing unknowns but ignore categorizing the known classes.
In outlier detection or one-/few-show learning \cite{fewshot}, samples of novel classes appear in training. 
In zero-shot learning \cite{zero}, semantic information from novel classes could be accessed. Such as zebra, an unknown class, could be identified given the idea that they are stripped horses, and abundant samples of horse and stripe patterns. 
Differently, OSR knows nothing about novel classes and should have high classification accuracy of the known meanwhile recognize unknowns, as illustrated in Fig. \ref{fig:intro} a). Due to limited space, some reviews \cite{review1,review2,review3} are recommended for more comprehensive conceptual distinctions. 

Most OSR researches focus on natural images, while medical OSR is still in its infancy. In medical fields, representative work like T3PO \cite{t3po} introduces an extra task to predict the input image augmentation, and samples with low probabilities are regarded as unknowns. 
CSL \cite{csl} uses generative adversarial neural networks (GAN) to generate proxy images and unknown anchors. 
As for natural images, a line of work tries to simulate unknowns using generated adversarial or counterfactual samples using GAN \cite{ganc,ganzero,gan_uncertain,ganopen}. However, whether unknown patterns could be generated by learning from the known is unclear. 
Some works learn descriptive feature representations. They enhance better feature separation between unknowns and knowns or assume the known features following certain distributions so that samples away from distributional centers could be recognized as unknowns \cite{rpl,place,podn,iiloss,clu,pmal,lslt}. 
Differently, this work categorizes densely distributed known features and recognizes sparse embedding space as unknowns, regardless of the specific distribution.

This work tackles OSR under the assumption that known features could be assembled compactly in feature embedding space, and remaining sparse regions could be recognized as unknowns. 
Inspired by this, the Open Margin Cosine Loss (OMCL) is proposed merging two components, Margin Loss with Adaptive Scale (MLAS) and Open-Space Suppression (OSS). 
The former enhances known feature compactness and the latter recognizes sparse feature space as unknown. 
Specifically, MLAS introduces the angular margin to the loss function, which reinforces the intra-class compactness and inter-class separability. Besides, a learnable scaling factor is proposed to enhance the generalization capacity. 
OSS generates feature space descriptors that scatter across a bounded feature space. By categorizing them as unknowns, it opens a classifier by recognizing sparse feature space as unknowns and suppressing the overconfidence of the known. 
An embedding space example is demonstrated in Fig. \ref{fig:intro} b), showing OMCL learns more descriptive features and more distinguishing known-unknown separation.

Considering medical OSR is still a nascent field, besides OMCL, we also proposed two publicly available benchmark datasets. One is microscopic images of blood cells, and the other is optical coherence tomography (OCT) of the eye fundus. OMCL shows good adaptability to different image modalities.

Our contributions are summarized as follows. 
\textbf{Firstly}, we propose a novel approach, OMCL for OSR in medical diagnosis. It reinforces intra-class compactness and inter-class separability, and meanwhile recognizes sparse feature space as unknowns.
\textbf{Secondly}, an adaptive scaling factor is proposed to enhance the generalization capacity of OMCL. 
\textbf{Thirdly}, two benchmark datasets are proposed for OSR. Extensive ablation experiments and feature visualization demonstrate the effectiveness of each design. The superiority over state-of-the-art methods indicates the effectiveness of our method and the adaptability of OMCL on different image modalities.

\section{Method}

In Section \ref{pre}, the open set problem and the formation of cosine Softmax are introduced. The two mechanisms MLAS and OSS are sequentially elaborated in Section \ref{MLAS} and \ref{OSS}, followed by the overall formation of OMCL in Section \ref{OMCL}.

\subsection{Preliminaries \label{pre}}
\textit{\textbf{Problem setting}}: 
Both closed set and open set classifiers learn from the training set $\mathcal{D}_{train}=\left\{(\boldsymbol{x}_i, y_i)\right\}_{i=1}^{N}$ with $N$ image-label pairs $(\boldsymbol{x}_i, y_i)$, where $y_i\in\mathcal{Y}=\left\{1, 2, ..., C\right\}$ is a class label. 
In testing, closed set testing data $\mathcal{D}_{test}$ shares the same label space $\mathcal{Y}$ with the training data. However, in the open set problem, unseen class $y_i=C+1$ could appear in testing \emph{i.e.} $y_i\in\mathcal{Y}_{open}=\left\{1, 2, ..., C, C+1\right\}$. 

\begin{figure}[t]
\includegraphics[width=\textwidth]{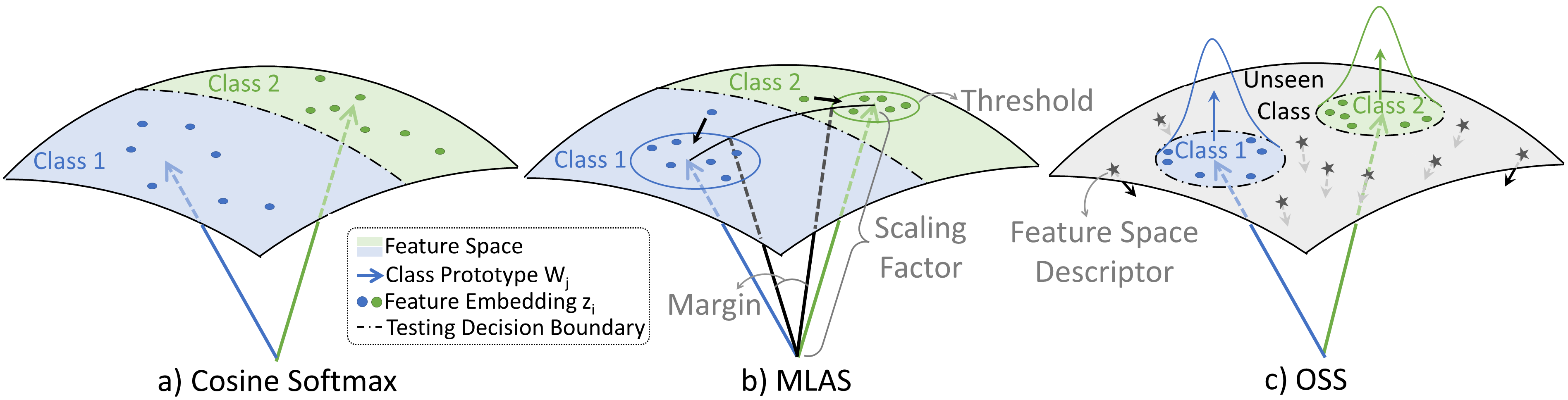}
\caption{Geometric interpretation of the cosine Softmax, MLAS, and OSS. MLAS introduces angular margin $m$ and threshold $t$ to reinforce the intra-class compactness and the inter-class separability, together with an adaptive scaling factor to enhance the adaptability. OSS opens a classifier by recognizing the sparse feature space as unknowns using the proposed feature space descriptors.} 
\label{fig:method}
\end{figure}

\textit{\textbf{Cosine Loss}}: 
The cosine Softmax is used as the basis of the OMCL. It transfers feature embeddings from the Euclidian space to a hyperspherical one, where feature differences depend merely on their angular separation rather than spatial distance. 
Given an image $\boldsymbol{x}_i$, its vectorized feature embedding $\boldsymbol{z}_i$, and its label $y_i$, the derivation progress of the cosine Softmax is
\begin{equation}
    S_{cos}
    =\underbrace{\frac{e^{\boldsymbol{W}_{y_i}^T\boldsymbol{z}_i}}{\sum_{j=1}^{C}e^{\boldsymbol{W}_{j}^T\boldsymbol{z}_i}}}_{Conventioanl Form}
    =\frac{e^{\parallel\boldsymbol{W}_{y_i}\parallel\parallel\boldsymbol{z}_i\parallel cos(\theta_{y_i,i})}}{\sum_{j=1}^{C}e^{\parallel\boldsymbol{W}_{j}\parallel \parallel\boldsymbol{z}_i\parallel cos(\theta_{j,i})}}
    =\underbrace{\frac{e^{s\cdot cos(\theta_{y_i,i})}}{\sum_{j=1}^{C}e^{s\cdot cos(\theta_{j,i})}}}_{Cosine Form}
\end{equation}
\noindent , where $\boldsymbol{W}_j$ denotes the weights of the last fully-connected layer (bias is set to 0 for simplicity). $\parallel\boldsymbol{W}_j\parallel=1$ and $\parallel\boldsymbol{z}_i\parallel=s$ are manually fixed to constant numbers 1 and $s$ by L2 normalization. $s$ is named the scaling factor. $cos(\theta_{j,i})$ denotes the angle between $\boldsymbol{W}_j$ and $\boldsymbol{z}_i$. By doing so, the direction of ${W}_j$ could be regarded as the prototypical direction of class $j$ as shown in Fig. \ref{fig:method} a). Samples with large angular differences from their corresponding prototype will be punished and meanwhile class-wise prototypes will be pushed apart in the angular space. 

Compared with Softmax, the cosine form has a more explicit geometric interpretation, promotes more stabilized weights updating, and learns more discriminative embeddings \cite{cosface,sphereface,negative}. 
Moreover, the L2 normalization constrains features to a bounded feature space, which allows us to generate feature space descriptors for opening a classifier (will be further discussed in Section \ref{OSS}).

\subsection{Margin Loss with Adaptive Scale (MLAS) \label{MLAS}}
MLAS serves three purposes. 
1) By applying angular margin, the intra-class compactness and the inter-class separability are strengthened. 
2) The threshold could represent the potential probability of the unknowns, which not only prepares for the open set but also learns more confident probabilities of the knowns.
3) A trainable scaling factor is designed to strengthen the generalization capacity. 
MLAS is:

\begin{equation}
\label{eq:mlas}
    S_{MLAS}
    =\frac{e^{s\cdot(cos(\theta_{y_i,i})-m)}}{e^{s\cdot(cos(\theta_{y_i,i})-m)}+e^{s\cdot t}+\sum_{j=1, j\neq y_i}^{C}e^{s\cdot cos(\theta_{j,i})}}
\end{equation}
$m$, $t$, and $s$ respectively denote margin, threshold, and learnable scaling factor, with corresponding geometric interpretation demonstrated in Fig. \ref{fig:method} b).

By using the angular margin, the decision boundary could be more stringent. 
Without it, the decision boundary is $cos(\theta_{1,i})>cos(\theta_{2,i})$ for the $i$-th sample of class 1. 
It becomes $cos(\theta_{1,i})>cos(\theta_{2,i})+m$ when using the margin, which leads to stronger intra-class compactness. Moreover, the angular similarities with other classes are punished in the denominator to increase inter-class separability. 

The threshold $t$ could be regarded as an extra dimension that prepares for unknown classes. Given the conventional input of Softmax as $[q_i^1, q_i^2, ..., q_i^C]\in \mathbb{R}^C$, ours could be understood as $[q_i^1, q_i^2, ..., q_i^C, t]\in \mathbb{R}^{C+1}$. Since $t$ is added, the class-wise output $q_i^c$ before Softmax is forced to have a higher value to avoid misclassification (at least larger than $t$). It reinforces more stringent learning and hence increases the feature compactness in the hyperspherical space.

A large $s$ makes the distribution more uniform, and a small $s$ makes it collapses to a point mass. 
In this work, it is learnable, with a learning rate 0.1$\times$ the learning rate of the model. It theoretically offers stronger generalization capacity to various datasets and is experimentally observed to converge to different values in different data trails and could boost performances.

LMCL \cite{cosface} and NMCL \cite{negative} are the most similar arts to ours. Differently, from the task perspective, these designs are proposed for closed-world problems. From the method perspective, an OSS mechanism is designed to tackle OSR leveraging generate pseudo-unknown features for discriminative learning. Moreover, an adaptive scaling factor is introduced for increasing generalization.

\subsection{Open-Space Suppression (OSS) \label{OSS}}
OSS generates feature space descriptors of bounded feature space. By categorizing them into an extra $C+1$ class, samples in sparse feature space could be recognized as unknown and the overconfidence of the known is suppressed.

OSS selects points scattered over the entire feature space, named descriptors, representing pseudo-unknown samples. Different from existing arts that generate pseudo-unknowns by learning from known samples, the OSS selects points scattered over the feature space. It guarantees all space could be possibly considered for simulating the potential unknowns. By competing with the known features, feature space with densely distributed samples is classified as the known, and the sparse space, represented by the descriptors, will be recognized as unknown. 

In this work, the corresponding descriptor set, with $M$ samples, is $\mathcal{D}_{desc}=\left\{(\boldsymbol{z}_i, C+1)\right\}_{i=1}^{M}$, where $\boldsymbol{z}_i \in \mathbb{U}[-s,s]^{d}$ subject to $\parallel \boldsymbol{z}_i\parallel=s$. $\mathbb{U}[-s,s]$ denotes random continuous uniform distribution ranges between $-s$ to $s$, and $d$ is the dimension of feature embeddings.
$s$ is trainable and the descriptors are dynamically generated with the training.
Fig. \ref{fig:method} c) demonstrates the geometric interpretation. During training, descriptors are concatenated with the training samples at the input of the last fully-connected layer, to equip the last layer with the discrimination capacity of known and unknown samples. The OSS is 
\begin{equation}
    S_{OSS}=\frac{e^{s\cdot t}}{e^{s\cdot t}+\sum_{j=1}^{C}e^{s\cdot cos(\theta_{j,i})}}
\end{equation}
where $t$ and $s$ follow the same definition in MLAS. 

Most similar arts like AL \cite{al} attempts to reduce misclassification by abandoning ambiguous training images. Differently, we focus on OSR and exploit a novel discriminative loss with feature-level descriptors for OSR.

\subsection{Open Margin Cosine Loss (OMCL) \label{OMCL}}
OMCL unifies MLAS and OSS into one formula, which is 
\begin{equation}
\label{eq:all}
    L_{OMCL}=-\frac{1}{N+M}\sum_{i=1}^{N+M} 
    \mathbb{I}_i log(S_{cos}) + \lambda\mathbb{I}_i log(S_{MLAS}) +\lambda\mathbb{I}_i log(S_{OSS})
\end{equation}
$\mathbb{I}_i$ equals 1 if the $i$-th sample is training data, and equals 0 if it belongs to the feature space descriptors. $\lambda$ is a weight factor. Since the output of the channel $C+1$ is fixed as $t$, no extra weights $\boldsymbol{W}_{C+1}$ are trained in the last fully-connected layer. As a result, OMCL does not increase the number of trainable weights in a neural network. During testing, just as in other works \cite{rpl,arpl}, the maximum probability of known classes is taken as the index of unknowns, where a lower known probability indicates a high possibility of unknowns.

\section{Result}
\subsection{Datasets, Evaluation Metrics, and Implementation Details}
Two datasets are adapted as new benchmarks for evaluating the OSR problem. Following protocols in natural images \cite{arpl,oscr}, half of the classes are selected as known and reminders as unknowns. Since the grouping affects the results, it is randomly repeated $K$ times, leading to $K$ independent data trials. The average results of $K$ trials are used for evaluation. The specific groupings are listed in the supplementary material, so that future works could follow it for fair comparisons. 

\textit{\textbf{BloodMnist}} contains 8 kinds of individual normal cells with 17,092 images \cite{blood}. Our setting is based on the closed set split and prepossessing from \cite{medmnistv2}. Classes are selected 5 rounds ($K$=5). In each trial, images belonging to 4 chosen classes are selected for training and closed-set evaluation. Images belonging to the other 4 classes in testing data are used for open set evaluation. 

\textit{\textbf{OCTMnist}} has 109,309 optical coherence tomography (OCT) images \cite{oct}, preprocessed following \cite{medmnistv2}. Among the 4 classes, 1 is healthy and the other 3 are retinal diseases. In data trail splitting, the healthy class is always in the known set, which is consistent with real circumstances, and trails equal to 3 ($K$=3). 

\textit{\textbf{Metrics}}: Following previous arts \cite{arpl,dias}, accuracy (ACC$_{c}$) validates closed set classification. Area Under the Receiver Operating Characteristic (AUROC$_{o}$), a threshold-independent value, measures the open set performances. Open Set Classification Rate (OSCR$_{o}$) \cite{oscr}, considers both open set recognition and closed set accuracy, where a larger OSCR indicates better performance.

\textit{\textbf{Implementation Details}}:
The classification network is ResNet18 \cite{resnet}, optimized by Adam with an initial learning rate of 1e-3 and a batch size 64. The number of training epochs is 200 and 100 for BloodMnist and OCTMnist respectively because the number of training samples in BloodMnist is smaller. Margin $m$, threshold $t$, $\lambda$ are experimentally set to -0.1, 0.1, and 0.5 respectively. Images are augmented by random crop, random horizontal flip, and normalization.

\begin{table}[t]
    \renewcommand{\arraystretch}{1.3}
    \caption{Comparison with state-of-the-art methods. The average of multiple trials is reported.}
    \label{tab:sota}
    \centering
    \begin{tabular}{c|ccc|ccc}
    \hline
    \multirow{2}{*}{Method (Pub'Year)}  &\multicolumn{3}{c|}{BloodMnist $K$=5} & \multicolumn{3}{c}{OCTMnist $K$=3}\\ 
    & Acc$_{c}\%$ & AUROC$_{o}\%$ & OSCR$_{o}\%$ & Acc$_{c}\%$ & AUROC$_{o}\%$ & OSCR$_{o}\%$ \\
    \hline
    Baseline~\cite{baseline} (ICLR'17) & 98.0 & 84.3 & 84.4 & 94.3 & 64.1 & 62.8 \\
    GCPL~\cite{gcpl} (CVPR'18)         & 98.1 & 85.5 & 85.0 & 94.8 & 65.5 & 64.2 \\
    RPL~\cite{rpl} (ECCV'20)           & 98.0 & 86.8 & 86.3 & 93.7 & 65.9 & 64.2 \\
    ARPL+CS~\cite{arpl} (TPAMI'21)     & \textbf{98.5} & 87.6 & 87.1 & 95.9 & 77.7 & 75.8 \\
    DIAS~\cite{dias} (ECCV'22)         & 98.4 & 86.3 & 85.7 & 96.0 & 74.1 & 72.5 \\
    OMCL (Ours) & 98.3 & \textbf{88.6} & \textbf{88.0} & \textbf{96.8} & \textbf{78.9} & \textbf{77.8} \\
    \hline
    \end{tabular}
\end{table}

\begin{table}[t]
    \renewcommand{\arraystretch}{1.3}
    \caption{Ablation studies on the effectiveness of MLAS and OSS in OMCL. Each result is the average of 5 trials on BloodMnist dataset.}
    \label{tab:ab}
    \centering
    \begin{tabular}{cc|ccc}
    \hline
    MLAS & OSS & Acc$_{c}\%$ & AUROC$_{o}\%$ & OSCR$_{o}\%$ \\
    \hline
      &   & 98.3 & 84.7 & 84.2 \\
    $\checkmark$ &   & 98.3 & 85.4 & 84.9 \\
      & $\checkmark$ & 98.3 & 86.8 & 86.3 \\
    $\checkmark$ & $\checkmark$ & 98.3 & \textbf{88.6} & \textbf{88.0} \\
    \hline
    \end{tabular}
\end{table}

\subsection{Comparison with State-of-the-art Methods}
As demonstrated in Table \ref{tab:sota}, the proposed OMCL surpasses state-of-the-art models, including typical discriminative methods, baseline\cite{baseline}, GCPL\cite{gcpl}, and RPL\cite{rpl}; latest generative model DIAS\cite{dias}; and ARPL+CS\cite{arpl} that hybrids both. All methods are implemented based on their official codes. Their best results after hyperparameter finetunes are reported. Results show the OMCL maintains the accuracy, meanwhile could effectively recognize unknowns.

\begin{figure}[t]
\includegraphics[width=\textwidth]{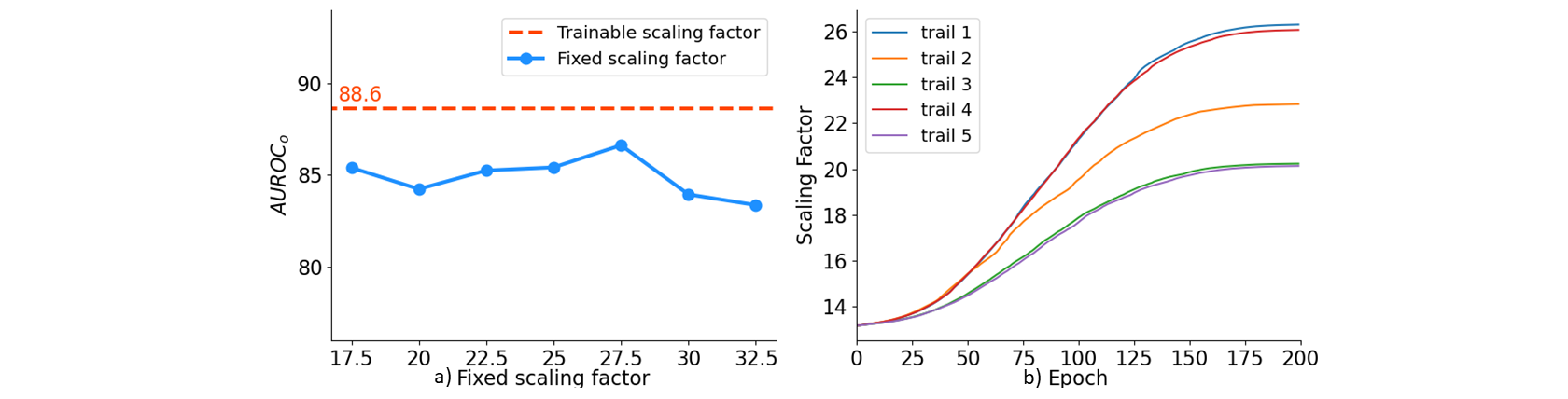}
\caption{Ablation studies of adaptive scaling factor on BloodMnist dataset.} 
\label{fig:ab_s}
\end{figure}

\begin{figure}[t]
\includegraphics[width=\textwidth]{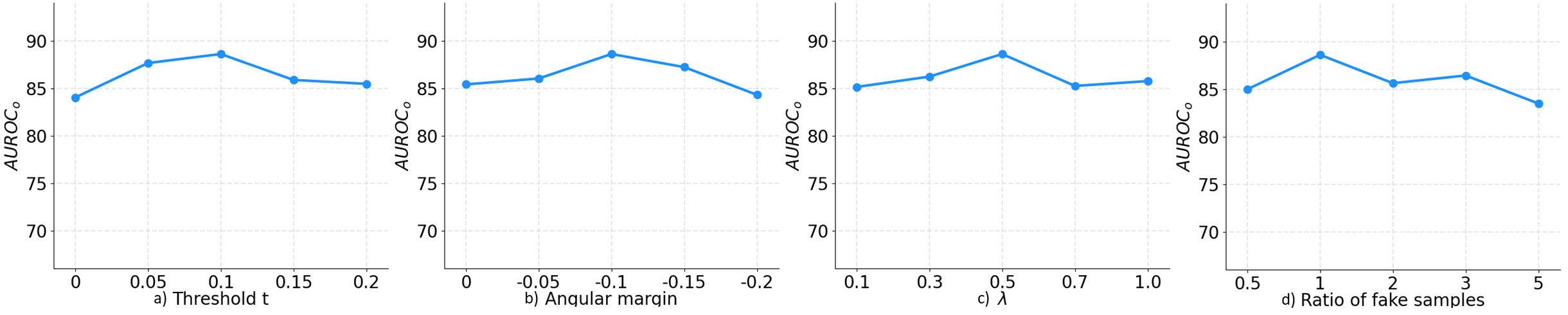}
\caption{Ablation studies of hyperparameters. Each result is the average of 5 trails on the BloodMnist dataset.} 
\label{fig:ab_all}
\end{figure}

\subsection{Ablation Studies}

\textit{\textbf{Effectiveness of MLAS and OSS}}: Table \ref{tab:ab} demonstrates the respective contributions of MLAS and OSS in OMCL. Each of them enhances the performances and they could work complementarily to further improve performances.

\textit{\textbf{Ablation Study of Adaptive Scaling Factor}}: Fig. \ref{fig:ab_s} a) demonstrates the effectiveness of the adaptive scaling factor. Quantitatively, the adaptive design surpasses a fixed one. Moreover, Fig. \ref{fig:ab_s} b) displays the scaling factor will converge to different values in different training trials. Both results demonstrate the effectiveness and the generalization capacity of the adaptive design.

\textit{\textbf{Ablation Study of Hyperparameters $t$, $m$, and $\lambda$}}: Fig. \ref{fig:ab_all} a), b), and c) respectively show the influence on results when using different hyperparameters. $t$ and $m$ are the threshold and angular margin, presented in equation \ref{eq:mlas}, and $\lambda$ is the trade-off parameter in equation \ref{eq:all} . 

\textit{\textbf{Ablation Study of}} $M$: Fig. \ref{fig:ab_all} d) illustrates the effect of the number of feature space descriptors upon results. The ratio 1:1 is experimentally validated as a proper ratio. Because a randomly generated descriptor could be extremely close to a known feature point, but classified as a novel category, which may disturb the training. If the number of descriptors is far more than that of the training samples (the 5 times shown in Fig. \ref{fig:ab_all} 4), the performance gets lower.

\textit{\textbf{Feature Visualization}}: Fig. \ref{fig:intro} b) visualizes the t-SNE results of features $\boldsymbol{z}$ of both known and unknown classes after dimension reduction. For each class, 200 samples are visualized and the perplexity of the t-SNE is set to 30. It shows that OMCL could learn better intra-class compactness and inter-class separability. Moreover, samples of unknown classes tend to be pushed away from known classes, incidcating the effectiveness of our designs.

\section{Conclusion}
In this paper, two publicly available benchmark datasets are proposed for evaluating the OSR problem in medical fields. Besides, a novel method called OMCL is proposed, under the assumption that 
known features could be assembled compactly in feature space and the sparse regions could be recognized as unknowns. 
The OMCL unifies two mechanisms, MLAS and OSS, into a unified formula. The former reinforces intra-class compactness and inter-class separability of samples in the hyperspherical feature space, and an adaptive scaling factor is proposed to empower the generalization capability. 
The latter opens a classifier by categorizing sparse regions as unknown using feature space descriptors.
Extensive ablation experiments and feature visualization demonstrate the effectiveness of each design. Compared to recent state-of-the-art methods, the proposed OMCL performs superior, measured by ACC, AUROC, and OSCR.


%
%
%
\bibliographystyle{splncs04}
\bibliography{refbib}

\end{document}